\documentclass[11pt]{article}

\usepackage[preprint]{acl}

\usepackage{times}

\usepackage[T1]{fontenc}

\usepackage[utf8]{inputenc}

\usepackage{microtype}

\usepackage{inconsolata}

\usepackage{graphicx}
\usepackage{multirow}

\usepackage[framemethod=TikZ]{mdframed}
\newmdenv[
  topline=false,
  bottomline=false,
  rightline=false,
  skipabove=\topsep,
  skipbelow=\topsep,
  leftmargin=8pt,
  rightmargin=0pt,
  innertopmargin=0pt,
  innerbottommargin=0pt,
  linewidth=2pt,
  linecolor=gray,
]{leftlinequote}

\usepackage{xcolor}
\usepackage{amsmath}
\definecolor{darkblue}{RGB}{0,0,139}
\definecolor{darkgreen}{RGB}{0,100,0}
\usepackage{booktabs}
\usepackage{tabularx}

\usepackage{array}                                    
\usepackage[table]{xcolor}                            
\usepackage{colortbl}                                 
\usepackage{placeins}                                 
\usepackage{float}                                    
\usepackage{multicol}                                 
\usepackage{multirow}                                 
\usepackage{amssymb}
\definecolor{promptbg}{rgb}{0.95,0.95,0.95} 

\usepackage{enumitem}

\usepackage{tikz}
\def\checkmark{\tikz\fill[scale=0.4](0,.35) -- (.25,0) -- (1,.7) -- (.25,.15) -- cycle;} 

\definecolor{ctxblue}{HTML}{004BFF}       
\definecolor{candorange}{HTML}{E85D04}    
\definecolor{goldgreen}{HTML}{00A86B}     
\definecolor{targetpurple}{HTML}{7B2CBF}  
\definecolor{datasetbg}{HTML}{FAFAFA} 
\definecolor{clsbg}{HTML}{DDEAFF}     
\definecolor{sftbg}{HTML}{FFD5CC}     
\definecolor{contbg}{HTML}{FFEAA3}    
\definecolor{headerbg}{HTML}{E9ECEF}  
\definecolor{rulegray}{HTML}{AAB2BD}

\newcommand{\ctx}[1]{\textcolor{ctxblue}{#1}}
\newcommand{\cand}[1]{\textcolor{candorange}{#1}}
\newcommand{\gold}[1]{\textcolor{goldgreen}{\textbf{#1}}}
\newcommand{\target}[1]{\textcolor{targetpurple}{\textbf{#1}}}

\newcolumntype{D}{>{\columncolor{datasetbg}\raggedright\arraybackslash}p{0.09\textwidth}}
\newcolumntype{C}{>{\columncolor{clsbg}\raggedright\arraybackslash}X}
\newcolumntype{S}{>{\columncolor{sftbg}\raggedright\arraybackslash}X}
\newcolumntype{T}{>{\columncolor{contbg}\raggedright\arraybackslash}X}

\title{Punching Above Their Weight: Classification-Head Fine-Tuning of Tiny Language Models (TLMs) for Verifiable Multiple-Choice Tasks}

\author{Bhavesh Sood \\
  Carnegie Mellon University \\
  \texttt{bsood@andrew.cmu.edu} \\\And
  Jaromir Savelka \\
  Carnegie Mellon University  \\
  \texttt{jsavelka@andrew.cmu.edu} \\}

\begin{document}
\maketitle
\begin{abstract}
We define \emph{Tiny Language Models} (TLMs) as models below roughly 3B parameters that fit on mainstream consumer devices. We study how to adapt them for and use them on verifiable multiple-choice tasks. We compare three LoRA-based fine-tuning paradigms (label generation, gold only, and our discriminative classification head) on a unified setup across several Qwen3 models from 0.6B to 8B and five benchmarks: HellaSwag, WinoGrande, PIQA, SciQ and ARC-C.
Classification-head fine-tuning reliably outperforms label generation ($+2$--$3$\%) at the 0.6B and 1.7B scales. Further, TLMs fine-tuned using the discriminative method are competitive to zero-/few-shot GPT-3 (175B), PaLM (540B) and GPT-4. The performance we report for Qwen3-0.6B and Qwen3-1.7B are SOTA on HellaSwag, WinoGrande, and PIQA.
\end{abstract}

\section{Introduction}
\begin{figure*}[t]
	\centering
	\includegraphics[width=\textwidth]{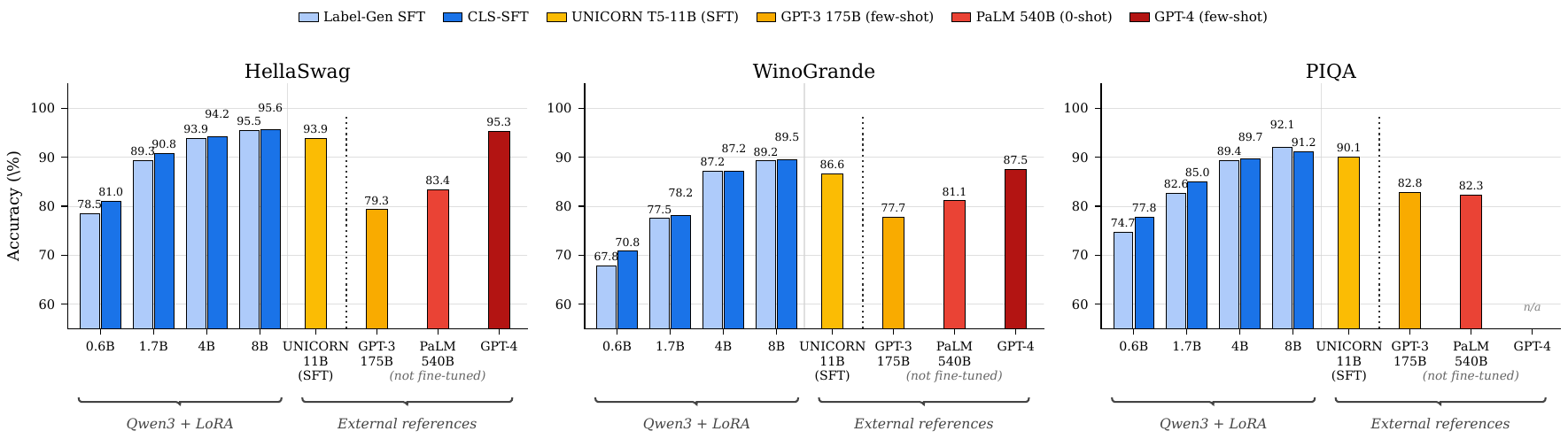}
	\caption{
		Accuracy on three commonsense benchmarks for Qwen3 + LoRA
		models fine-tuned with Label-Gen SFT and Classification-head SFT
		(CLS), compared to four external reference models: UNICORN T5-11B
		\citep{lourie2021unicorn}, GPT-3 175B \citep{brown2020gpt3}, PaLM
		540B \citep{chowdhery2022palm}, and GPT-4 \citep{openai2023gpt4}.
		UNICORN is the only fine-tuned external reference. PIQA is not
		reported in the GPT-4 Technical Report.
		We observe that at the 1.7B scale, fine-tuned Qwen3 already exceeds the reported GPT-3
		and PaLM 540B results on HellaSwag and PIQA, and exceeds GPT-3 on
		WinoGrande, despite being roughly $100\times$ smaller than GPT-3 and
		$320\times$ smaller than PaLM.
		}
	\label{fig:teaser}
\end{figure*}
Running large language models (LLMs) on common consumer hardware has become increasingly important as the capabilities and ubiquity of LLMs rise at unprecedented pace. Yet, vast majority of published research focuses on LLMs that cannot be effectively run on such devices. Hence, we argue that there is a great need to coin a novel notion of a Tiny LM (TLM) that is motivated by contemporary practical constraints stemming from common consumer hardware. 
Because TLMs remain understudied their deployments are often sub-optimal with lower accuracy than what they can deliver.

This study of fine-tuning paradigms and evaluation approaches for TLMs offers actionable guidance on how to adapt and use a TLM. The study is task-agnostic and only assumes a setting in which there is a finite set of candidate outputs for each data point with one of the candidates deemed as correct (e.g., classification, span annotation, or typical applications of LLM-as-a-judge). We denote such tasks as verifiable multiple-choice. Our specific focus is on commonsense reasoning tasks as these would likely be the main TLMs use case on local devices. However, such tasks have been shown to be challenging for LMs of small sizes \cite{li2025small,lu2024slm}.
\vspace{0.3em}

\noindent This study makes the following \textbf{contributions}:

\begin{enumerate}
	\item It defines and motivates the practical notion of \emph{Tiny Language Models (TLMs)}.
	\item It describes a fine-tuning approach based on discriminative classification head suitable for application of TLMs on multiple-choice tasks.
	\item It is the first comparison of the three fine-tuning paradigms across five well-established benchmarks focused on TLMs, and the reported performance on most of them is SOTA.
\end{enumerate}

The main \textbf{findings} of this study are: 
\begin{enumerate}[label=\alph*)]
	\item In TLMs, the fine-tuning based on classification head offers significantly better performance than approaches based on fine-tuning for next-token (label) prediction.
	\item Modifying seemingly mundane details of inference can result in dramatic discrepancies in observed performance of a TLM.
	\item In larger models ($>3$B), the gap is closed and there is no significant difference between the fine-tuning approaches.
	\item Additionally, the memorisation-intensive science benchmarks benefit less from the classification approach, but with no statistically significant loss either.
\end{enumerate}

 Our findings suggest that the previously reported performance of TLMs may be simply an artifact of how they are typically fine-tuned and utilized (often sub-optimally). This calls for careful rethinking of evaluation and progress in this space. While it is generally accepted that TLMs do not perform adequately on common reasoning tasks Figure~\ref{fig:teaser} suggests quite a different story.

\section{Tiny Language Models}
Weights of a 3B model in BF16/FP16 consume 6--7GB of memory (see Table \ref{tab:model-memory} or Appendix \ref{app:model-device-memory}). 
The dominant memory tiers of common consumer hardware such as iPhone 16/17 or an RTX~5060, sit at 8GB (see Table \ref{tab:consumer-device-memory1} or Appendix \ref{app:model-device-memory}).
Only LLMs with $\leq$3B parameters leave meaningful headroom for an operating system, KV cache, and activations at usable context lengths whereas a 4B model~($\sim$8GB weights) would effectively exhaust common consumer hardware. Therefore:

\vspace{8pt}

\begin{leftlinequote}
We define TLMs as models that can be effectively deployed on mainstream consumer devices, typically requiring less than 8 GB of memory at BF16/FP16 precision. This requirement is met by LMs with size below or around 3 billion parameters.
\end{leftlinequote}

\noindent Prior work defines Small Language Models (SLMs) with somewhat looser and unclear upperbound~\citep{lu2024slm}. We argue that there is a need for defining the tighter $\sim$3B parameters class (i.e., TLMs). This size is the maximum that effectively fits in standard BF16/FP16 precision on common consumer hardware that leaves room for normal device operation. This threshold aligns with the leading on-device deployments such as Apple Intelligence's foundation model~\citep{gunter2024apple,apple2025foundation} or Meta's Llama 3.2 lightweight tier~\citep{grattafiori2024llama,llama32}.

\begin{table}[t]
	\scriptsize
	\setlength{\tabcolsep}{4.5pt}
	\renewcommand{\arraystretch}{1.15}
	\begin{tabular}{lccc}
		\toprule
		\textbf{Model}                            & \textbf{FP32} & \textbf{BF16 / FP16} & \textbf{FP8} \\
		\midrule
		Qwen3-0.6B~\citep{qwen3}                  & 3.0GB         & 1.5GB                & 0.8GB        \\
		Llama-3.2-1B~\citep{llama32}              & 5.0GB         & 2.5GB                & 1.2GB        \\
		Qwen3-1.7B~\citep{qwen3}                  & 6.9GB         & 3.4GB                & 1.7GB        \\
		Llama-3.2-3B~\citep{llama32}              & 12.8GB        & 6.4GB                & 3.2GB        \\
		\midrule
		Qwen3-4B~\citep{qwen3}                    & 16.1GB        & 8.1GB                & 4.0GB        \\
		Llama-3.1-8B~\citep{grattafiori2024llama} & 32.1GB        & 16.1GB               & 8.0GB        \\
		Qwen3-8B~\citep{qwen3}                    & 32.8GB        & 16.4GB               & 8.2GB        \\
		\bottomrule
	\end{tabular}
	\captionof{table}{GPU memory requirements for model weights at each precision. Values are based on
	reported Hugging Face safetensor shard sizes.}
	\label{tab:model-memory}
\end{table}

\begin{table}[t]
	\scriptsize
	\renewcommand{\arraystretch}{1.15}
	\begin{tabular}{p{0.13\columnwidth}p{0.34\columnwidth}p{0.39\columnwidth}}
		\toprule
		\textbf{Memory} & \textbf{Smartphones}                                                              & \textbf{Laptops / GPUs} \\
		\midrule
		8GB             & iPhone 16 / 16 Plus / 16 Pro / 16 Pro Max; iPhone 17
		                & RTX 4060 / 4070; RTX 5050 / 5060 / 5070 (8GB)                                                               \\

		12GB            & iPhone 17 Pro / Pro Max / Air; Galaxy S26 Ultra (256GB / 512GB)
		                & RTX 4080; RTX 5070 (12GB) / 5070 Ti                                                                         \\
		\midrule

		16GB            & Pixel 10 Pro / Pro XL; Galaxy S26 Ultra (1TB)
		                & RTX 4090; RTX 5080; MacBook Air baseline (unified)                                                          \\

		16--32GB        & --
		                & MacBook Air configurations, 16--32GB unified                                                                \\

		24--32GB        & --
		                & RTX 5090 (24GB laptop / 32GB desktop)                                                                       \\

		$>$32GB         & --
		                & High-end MacBook Pro M-Max / M-Ultra, 64--128GB unified \textbf{(premium price tier only)}                           \\
		\bottomrule
	\end{tabular}
	\captionof{table}{Representative memory budgets of commodity consumer devices (2025--2026).}
	\label{tab:consumer-device-memory1}
\end{table}
Table \ref{tab:intro_zeroshot_vs_cls} shows that utilization of TLMs in zero-shot settings results in poor performance on our benchmarks (often close to random). Second, the table showcases that different viable approaches to evaluation (e.g., probability of the next token versus log-probability of the whole sequence) lead to dramatic gaps in performance. Third, fine-tuning TLMs for the specific tasks leads to superior performance ($\sim$+25\%) that can be on par with zero-/few-shot versions of frontier LLMs. These observations motivated us to systematically investigate several approaches to fine-tuning and evaluation in TLMs. The study provides rich basis for rethinking of evaluation and progress in this space.

\section{Related Work}

\paragraph{Defining TLMs in the SLM landscape.}
The boundary between \emph{Small Language Models} (SLMs) and larger
LLMs is not standardized. Surveys note that SLMs are usually motivated
by lower latency, reduced cost, customization, privacy, and deployment
in resource-constrained settings \citep{wang2024comprehensive_slm}.
However, the parameter ranges differ substantially across papers:
\citet{srivastava2025thinkslm} consider SLMs up to roughly 30B
parameters, \citet{lu2024smallmodels} focus on decoder-only models from
100M to 5B parameters, and \citet{gupta2025slmsurvey} study SLMs mainly
in the 1B--9B range. These definitions are all motivated by practical
deployment on consumer devices such as phones,
laptops, and edge hardware, where local execution can reduce cloud
dependence and improve privacy \citep{yuan2023mobilefirmware,
dubiel2024ondevice}. In this work, we use
the memory constraints of such devices as the main motivation for
drawing a tighter boundary.

\begin{table}
\footnotesize
\setlength{\tabcolsep}{4pt}
\renewcommand{\arraystretch}{1.1}
\resizebox{\columnwidth}{!}{%
\begin{tabular}{llccccc}
	\toprule
	\multirow{2}{*}{\textbf{Dataset}}        & \multirow{2}{*}{\textbf{Model}}
											& \multirow{2}{*}{\textbf{Random}}
											& \multicolumn{2}{c}{\textbf{Zero-shot}}
											& \multirow{2}{*}{\textsc{CLS SFT}}
											& \multirow{2}{*}{$\boldsymbol{\Delta}$}                                                                              \\ 
											&                                         &       & \textsc{Label-Gen} & \textsc{Log-Prob} &                &        \\
	\midrule
	HellaSwag                                & Qwen3-0.6B                              & 25.0  & 25.58               & 36.27             & \textbf{80.95} & +44.68 \\
	                                & Qwen3-1.7B                              & 25.0  & 59.42               & 44.58             & \textbf{90.81} & +31.39 \\
	                                & Qwen3-8B                                & 25.0  & 80.11               & 55.21             & \textbf{95.62} & +15.51 \\
	\midrule
	WinoGrande                               & Qwen3-0.6B                              & 50.0  & 49.58               & 55.97             & \textbf{70.80} & +14.83 \\
	                               & Qwen3-1.7B                              & 50.0  & 51.39               & 57.10             & \textbf{78.16} & +21.06 \\
	                               & Qwen3-8B                                & 50.0  & 61.01               & 65.42             & \textbf{89.53} & +24.11 \\
	\midrule
	PIQA                                     & Qwen3-0.6B                              & 50.0  & 51.47               & 67.41             & \textbf{77.80} & +10.39 \\
	                                     & Qwen3-1.7B                              & 50.0  & 71.60               & 72.47             & \textbf{85.04} & +12.57 \\
	                                     & Qwen3-8B                                & 50.0  & 86.51               & 76.28             & \textbf{91.19} & +4.68  \\
	\midrule
	SciQ                                     & Qwen3-0.6B                              & 25.0  & 49.20               & 67.60             & \textbf{90.70} & +23.10 \\
	                                     & Qwen3-1.7B                              & 25.0  & 88.30               & 81.90             & \textbf{94.20} & +5.90  \\
	                                     & Qwen3-8B                                & 25.0  & 95.20               & 90.30             & \textbf{96.00} & +0.80  \\
	\midrule
	ARC-Challenge                            & Qwen3-0.6B                              & 25.0  & 27.55               & 31.67             & \textbf{63.61} & +31.94 \\
	                            & Qwen3-1.7B                              & 25.0  & 73.13               & 39.91             & \textbf{79.57} & +6.44  \\
	                            & Qwen3-8B                                & 25.0  & 90.04               & 55.11             & \textbf{91.07} & +1.03  \\
	\midrule
	\multicolumn{2}{l}{\emph{Mean (0.6B)}} & 35.0                                    & 40.68 & 51.78               & 76.77             & \textbf{+25.0}          \\
	\multicolumn{2}{l}{\emph{Mean (1.7B)}} & 35.0                                    & 68.77 & 59.19               & 85.56             & \textbf{+15.5}          \\
	\multicolumn{2}{l}{\emph{Mean (8B)}}   & 35.0                                    & 82.55 & 68.46               & 92.68             & \textbf{+9.2}           \\
	\bottomrule
\end{tabular}%
}
\captionof{table}{
	Zero-shot accuracy of Qwen3 models under two inference
	protocols compared with our LoRA classification-head \textsc{Cls SFT}.
	\textsc{Label-Gen} uses next-token argmax over answer-letter tokens. \textsc{LogProb} selects
	the candidate with the highest sequence log-probability. 
	}
\label{tab:intro_zeroshot_vs_cls}
\end{table}

The term \emph{TinyLLM / TinyLM / TLM} is even less consistent. Recent edge-agent work
studies sub-3B SLMs for agentic tasks while using TinyLLM and SLM
terminology closely together \citep{haque2025tinyllmagentic}. Other
work uses the TinyLLM name for models as small as 30M--120M parameters
\citep{kandala2024tinyllm}, while ``tiny language model'' work also
appears at the 1B--1.5B scale \citep{tang2024pangu}. Related surveys of
tiny language models for automation and control do not provide a single parameter cutoff for TinyLLMs
\citep{lamaakal2025tinyautomation}. We therefore define
\textbf{TLMs} operationally as models below 3B parameters. This
cutoff separates them from the broader SLM category, matches common
open-weight size tiers, and reflects the practical goal that the model
weights should fit on many consumer devices with enough remaining memory for
normal execution. Appendix~\ref{app:model-device-memory} lists
representative model families and device memory capacities supporting
this choice in detail.

\paragraph{Prediction-interface sensitivity in multi-choice evaluation.}
Multi-choice evaluation of LLMs is highly sensitive to how the answer is
extracted from the model. \citet{lyu2024beyondprobabilities} show that
probability-based scoring and free-form generation often disagree on the
selected option, and \citet{molfese2025rightanswer} find that common
extraction protocols misalign with human judgement, especially when models
produce rationales before committing to a letter. Even tokenisation
details matter; \citet{sanzguerrero2025mindgap} report up to $11\%$
accuracy swings, and reshuffled model rankings, depending solely on
whether the space after ``Answer:'' is tokenised with or apart from the
answer letter. More broadly, \citet{polo2024prompteval} argue that
single-prompt scores are unstable and propose estimating a distribution
over prompts. These findings push us to rethink how progress is measured
on multi-choice benchmarks, and motivate our treatment of the
\emph{training-and-evaluation interface} as a first-class object. Rather
than treating instruction-formatted label generation, LM-head log-probability
scoring, and classification-head candidate scoring as interchangeable, we
evaluate them as distinct prediction strategies. As our results show, the
gap between them is substantial for TLMs, where switching from the trained
interface to a different scoring interface can materially change reported
accuracy.

\paragraph{Fine-tuning approaches for multi-choice tasks.}
The strongest published fine-tuned reference on the commonsense benchmarks
we consider is UNICORN \citep{lourie2021unicorn}, which adapts T5-11B
\citep{raffel2020t5} via multitask transfer on RAINBOW
($\alpha$NLI, CosmosQA, HellaSwag, PIQA, SocialIQa, WinoGrande) followed
by per-task fine-tuning, scoring candidates by generation likelihood at
test time, structurally similar to the log-probability scoring used by
\texttt{lm-evaluation-harness} \citep{eval-harness} and to our LP-Sum
metric. Our work differs along three axes. We use
decoder-only Qwen3 models $1.4$ to $19\times$ smaller than T5-11B, train
a single-task LoRA adapter per benchmark with no multitask transfer, and
replace generative candidate likelihood with a discriminative
classification head.
\citet{yousefiramandi2025finetuning} also study classification-head
fine-tuning for decoder-only LLMs, comparing an embedding-based classifier
against instruction-formatted label generation under LoRA/QLoRA, and show
that classifier can outperform label generation on text classification. Our
work is complementary but studies a different setting. Rather than assigning a label to a
single input document, we reformulate verifiable multiple-choice tasks
as candidate-sequence ranking

\section{Method}
\label{sec:method}
We study Classification-head SFT, a discriminative alternative to instruction-formatted label-generation supervised fine-tuning for multiple-choice benchmarks. Instead of fine-tuning
the model to generate an answer option in response to a prompt, we fine-tune
it to assign a scalar score to each candidate input. The prediction is then obtained by selecting the
highest-scoring candidate.

\subsection{Constructing Candidate Sequences}
\label{sec:method:format}

Our formulation 
applies to any task where each data point has a single correct output and a finite set of incorrect alternatives.
The central requirement is that each candidate and its context can be converted into an input sequence that can be
scored by the model. Thus, instead of fine-tuning the model to generate the
answer, we transform the task into a candidate-scoring problem.

Formally, for an input instance $x$ with candidate outputs
$\{y_1,\ldots,y_K\}$ and gold index $y_i$, we construct $K$ candidate
sequences:
\[
	s_k = f(x, y_k),
\]

\noindent where $f$ is a task-specific input transformation function. The same scoring model
is then applied to each $s_k$, producing one scalar score $z_k$
per candidate. This gives a unified fine-tuning format across tasks whose
surface forms may differ (e.g., sentence completion, fill-in-the-blank,
or question answering). This candidate-sequence framing connects directly to log-probability-based
  multiple-choice inference \citep{brown2020language,lyu2024beyondprobabilities},
  in which each candidate is scored independently. Our method differs in
  what does the scoring: a single linear scoring head over the final hidden
  state, not a per-token language-modeling head over the candidate's tokens.

The form of $f$ depends on the task structure: it concatenates context
  with candidate for completion tasks (HellaSwag), substitutes a candidate
  into a blank for fill-in-the-blank tasks (WinoGrande), and pairs a
  question with each candidate answer for QA tasks (PIQA, ARC-Challenge,
  SciQ). Table~\ref{tab:format_example_main} shows 
  $f$ on a HellaSwag example; the per-dataset templates are given in
  Appendix~\ref{app:input-formats}.

\begin{table*}[t]
\centering
\footnotesize
\setlength{\tabcolsep}{6pt}
\renewcommand{\arraystretch}{1.2}
\arrayrulecolor{rulegray}
\begin{tabularx}{\textwidth}{|C|S|T|}
\hline
\rowcolor{headerbg}
\textbf{\textsc{Cls}: candidate-sequence scoring} &
\textbf{Label-Gen SFT: answer-label generation} &
\textbf{Gold-Only SFT: gold-only continuation} \\
\hline

\textit{All candidates are scored independently:}\par\smallskip
A. \ctx{Roof shingle removal: A man is sitting on a roof. He} \cand{is using wrap to wrap a pair of skis.}\par
B. \ctx{Roof shingle removal: A man is sitting on a roof. He} \cand{is ripping level tiles off.}\par
C. \ctx{Roof shingle removal: A man is sitting on a roof. He} \cand{is holding a Rubik's cube.}\par
D. \ctx{Roof shingle removal: A man is sitting on a roof. He} \gold{starts pulling up roofing on a roof.}\par\smallskip
\textit{Training: cross-entropy over four scalar candidate scores.}
&
\texttt{<|im\_start|>user}\par
Choose the most plausible ending.\par\smallskip
Context: \ctx{Roof shingle removal: A man is sitting on a roof. He}\par\smallskip
A. is using wrap to wrap a pair of skis.\par
B. is ripping level tiles off.\par
C. is holding a Rubik's cube.\par
D. starts pulling up roofing on a roof.\par\smallskip
Answer with only one letter: A, B, C, or D.\par
Answer:\texttt{<|im\_end|>}\par
\texttt{<|im\_start|>assistant} \target{D}\par\smallskip
\textit{Training: next-token prediction on the answer label only.}
&
\textit{Only the gold continuation is used:}\par\smallskip
\ctx{Roof shingle removal: A man is sitting on a roof. He} \gold{starts pulling up roofing on a roof.}\par\smallskip
\textit{Training: next-token prediction on the gold ending tokens only; context tokens are masked from the loss. At evaluation, all candidate endings are scored by LM log-probability.}
\\
\hline
\end{tabularx}
\caption{
Example training formats for one HellaSwag instance. Blue text denotes the shared context, orange denotes incorrect candidate endings, green denotes the gold ending, and purple denotes the supervised answer label. The three columns correspond to our three training paradigms. \textsc{Cls} directly compares all candidate sequences, Label-Gen SFT trains the model to emit an answer label, and Gold-Only SFT trains only on the gold continuation.
}
\label{tab:format_example_main}
\end{table*}

\subsection{Classification head architecture}
\label{sec:method:arch}
We implement the candidate scoring model  
by loading the
pretrained decoder-only transformer backbone and replacing its
language-modeling head with a sequence-level scoring head.\footnote{We use the
\texttt{AutoModelForSequenceClassification} interface from Hugging Face
\citep{wolf2020transformers} with \texttt{num\_labels=1}.} Whereas the standard LM head is a linear projection
  $\mathbf{W}_{\!\mathrm{LM}}\in\mathbb{R}^{d\times |V|}$ that produces a
  distribution over the full vocabulary at every position, our scoring
  head is a bias-free linear projection $w\in\mathbb{R}^{d\times 1}$
  applied only to the final non-padding hidden state of the candidate
  sequence. 
The head contains no intermediate hidden layer and applies no non-linear
activation.

Formally, for a minibatch of $B$ examples with $K$ candidates per
example, we tokenize and pad all candidate sequences to length $L$,
obtaining tensors of shape $[B,K,L]$. We then flatten the candidate
dimension and run the model on the resulting batch of shape $[BK,L]$.
Let $h_{i,k} \in \mathbb{R}^{d}$ denote the final non-padding hidden
state for candidate $k$ of example $i$. The classification head computes
\begin{equation*}
	z_{i,k} = w^\top h_{i,k},
\end{equation*}
where $w \in \mathbb{R}^{d}$ is the learned scoring vector. For example,
Qwen3-0.6B has an embedding size of $d=1024$, so the classification head for it adds only 1024 trainable parameters
when \texttt{num\_labels=1}. 

\subsection{Fine-tuning Objective}
\label{sec:method:loss}

Given a gold answer index $y_i \in \{1,\ldots,K\}$ for example $i$, we
treat the $K$ candidate scores as logits over the answer choices. For a
minibatch of $B$ examples, the model produces a score matrix
$Z \in \mathbb{R}^{B \times K}$, where $z_{i,k}$ is the scalar score for
candidate $k$ of example $i$. We minimize the average cross-entropy loss
over the minibatch:
\begin{equation*}
	\mathcal{L}
	=
	- \frac{1}{B}
	\sum_{i=1}^{B}
	\log
	\frac{\exp(z_{i,y_i})}
	{\sum_{k=1}^{K} \exp(z_{i,k})}.
\end{equation*}

\noindent This objective trains the model discriminatively over the candidates for
the same example. The softmax is
therefore applied only across the $K$ candidate scores during the loss
computation, not inside the model itself.
In logit space, the gradient raises the score assigned
to the gold candidate and suppresses the scores of incorrect
candidates. This differs from Label-Gen SFT, where the model is
fine-tuned to generate an answer label or answer tokens given a
prompt. In our formulation, the candidate texts themselves are directly
compared through their learned sequence-level scores.

\section{Experiments}
\label{sec:experiments}

\subsection{Datasets}
\label{sec:datasets}
Our experiments use five English multi-choice benchmarks grouped into
two tiers. The three \textbf{primary commonsense reasoning benchmarks}: HellaSwag, WinoGrande, and PIQA are widely reported in
language-model evaluations~\citep{brown2020language,grattafiori2024llama}
and remain non-trivial even for strong models. Together they cover a wide range of complementary
forms of commonsense reasoning (event continuation, coreference
resolution, physical commonsense). These benchmarks also reflect suitable use cases for on-device TLMs such as choosing the next likely action in an assistant
workflow or resolving what a short message refers to.
The additional two \textbf{science reasoning
benchmarks}: SciQ and ARC-Challenge are memorisation-intensive
question-answering tasks that depend more on stored factual knowledge
than on candidate discrimination. We include them to test whether
the candidate-scoring advantage transfers beyond commonsense reasoning.

\begin{table}
	\centering
	\scriptsize
	\setlength{\tabcolsep}{3pt}
	\renewcommand{\arraystretch}{1.12}
	\begin{tabularx}{\columnwidth}{@{}lXcrr@{}}
		\toprule
		\textbf{Dataset} & \textbf{Type} & \textbf{Choices} & \textbf{Train} & \textbf{Eval} \\
		\midrule
		\multicolumn{5}{@{}l}{\textit{Primary commonsense reasoning benchmarks}}                       \\
		HellaSwag        & Completion    & 4                & 39{,}905       & 10{,}042      \\
		WinoGrande (xl)  & Fill-in-the-blank & 2                & 40{,}398       & 1{,}767       \\
		PIQA             & Physical QA   & 2                & 16{,}113       & 1{,}838       \\
		\midrule
		\multicolumn{5}{@{}l}{\textit{Memorization intensive science benchmarks}}                        \\
		SciQ             & Science QA    & 4                & 11{,}679       & 1{,}000       \\
		ARC-Chal         & Science QA    & 4                & 1{,}119        & 1{,}172       \\
		\bottomrule
	\end{tabularx}
	\caption{
		Dataset statistics. Note that ARC-Challenge has a significantly smaller training set (1.1k) than the other four benchmarks.
	}
	\label{tab:dataset_stats}
\end{table}

\paragraph{HellaSwag} \citep{zellers2019hellaswag} is a 4-way
sentence-completion task with adversarially filtered distractors that
remain locally fluent and grammatically compatible with the context, so that simple surface cues are insufficient and
commonsense event understanding is required. The model must decide which continuation is coherent with
the situation described by the context.

\paragraph{WinoGrande} \citep{sakaguchi2020winogrande} is a 2-way
fill-in-the-blank coreference task at scale, adversarially filtered to
suppress annotation artifacts and word-association shortcuts from pretraining. We use the \texttt{winogrande\_xl} version. Correct
resolution typically depends on implicit commonsense understanding rather than
syntax alone.
\paragraph{PIQA} \citep{bisk2020piqa} is a 2-way physical-commonsense
dataset in which two lexically similar candidate solutions are paired with
an everyday goal. Solving it requires reasoning about object
affordances, materials, and practical physical constraints in the physical world.

\paragraph{SciQ} \citep{welbl2017crowdsourcing} is a 4-way
crowdsourced science multiple-choice benchmark spanning physics,
chemistry, biology, and earth science. We use it without the support passages.

\paragraph{ARC-Challenge} \citep{clark2018think} is a 4-way
grade-school science question-answering benchmark, restricted to the
``Challenge'' set of items that neither a retrieval baseline nor a
word co-occurrence baseline can solve.


\subsection{Models}
\label{sec:models}

We evaluate four open-weight decoder-only models from the
Qwen3 family \citep{qwen3} spanning \textbf{0.6B}, \textbf{1.7B},
\textbf{4B}, and \textbf{8B} parameters. The first two fall within
our TLM regime ($\leq$3B parameters), while the 4B and 8B models
let us test whether the same trends persist beyond it. This set
provides a controlled scaling axis within a recent, competitive
open-weight family.

All models are loaded and fine-tuned in \texttt{bfloat16}. We disable
Qwen3's thinking mode throughout all experiments so that models are
fine-tuned and evaluated to produce the final answer, without
generating intermediate reasoning. For Label-Gen SFT, we
use the default Hugging Face chat template associated with each model.

\subsection{Experimental Design}
\label{sec:expdesign}

For HellaSwag and PIQA, we use the public validation split as the 
evaluation set because test labels are private, otherwise we use the available test split.
All the datasets share a common verifiable structure:
each example contains a finite set of candidate answers and exactly one
correct choice. With the candidate-sequence formulation introduced in
Section~\ref{sec:method:format}, they provide a meaningful testbed for
studying fine-tuning strategies for TLMs.

For each dataset and model, we compare three fine-tuning paradigms
under a unified training configuration so that the only difference
between conditions is the fine-tuning objective.

\paragraph{Fine-tuning Objectives.}
We evaluate the following three paradigms.

\begin{itemize}\itemsep0pt
	\item \textbf{\textsc{Cls}: Classification-head SFT.}
	      This is our proposed method. Each candidate sequence is scored by a
	      sequence-level classification head, and the model is trained with
	      cross-entropy over the candidate scores.

	\item \textbf{Label-Gen SFT: Instruction-formatted label-generation supervised fine-tuning.}
	      The model receives a chat-style multiple-choice prompt containing
	      all answer options and is trained to generate the correct answer
	      letter, such as `` A'' or `` B'' \citep{sanzguerrero2025mindgap}. The loss is applied only to the
	      target answer token.

	\item \textbf{Gold-Only SFT: Gold-only continuation supervised fine-tuning.}
	      The model is trained only on the gold context-candidate sequence in the same format as \textsc{Cls-SFT}
	      using the standard next-token language-modeling objective, with
	      context tokens masked from the loss. This tests whether
	      increasing the likelihood of the correct continuation is
	      sufficient, without explicitly training on incorrect candidates.
\end{itemize}

\paragraph{Training setup.} All three paradigms train LoRA adapters
\citep{hu2022lora} on a frozen Qwen3 backbone for 2 epochs at
learning rate $2 \times 10^{-4}$, with per-device batch size 8,
gradient accumulation over 8 steps. Each fine-tuning run (with evals)
took roughly 12--16 hours on a single NVIDIA L4 GPU. Full LoRA and
optimizer hyperparameters are in
Appendix~\ref{app:lora_hyperparams}.

\paragraph{Evaluation protocols.}
\noindent We evaluate each training paradigm in its native prediction
interface: \textsc{Cls} uses classification-head argmax over candidate
scores, Label-Gen SFT uses next-token answer-label accuracy, and
Gold-Only SFT uses LM log-probability scoring over candidate
continuations. We also evaluate the Label-Gen SFT adapter as a candidate
scorer by applying LM log-probability scoring to the same
candidate sequences used by \textsc{Cls}. We denote this diagnostic
evaluation as Label-Gen LogProb. Additional diagnostics
including the LP-Full/LP-Sum/LP-Avg/LP-Char metrics, are described in Appendix~\ref{app:evaluation_protocols}.


\begin{table*}[t]
  \centering
  \footnotesize
  \setlength{\tabcolsep}{4pt}
  \renewcommand{\arraystretch}{1.10}
  \begin{tabular}{lclcccccc}
    \toprule
    \textbf{Dataset} & \textbf{TLM} & 
	\textbf{\shortstack[c]{Qwen3\\Params}} &
    \textbf{\textsc{Cls-SFT}} &
    \textbf{\shortstack[c]{Label-Gen\\SFT}} &
    \textbf{$\boldsymbol{\Delta}$} &
    \textbf{\shortstack[c]{Label-Gen\\LogProb}} &
    \textbf{\shortstack[c]{Gold-Only\\SFT}} &
    \textbf{$p$} \\
    \midrule

    \multirow{4}{*}{HellaSwag}
      &\checkmark & 0.6B & \textbf{80.95} & 78.50          & $+2.45$ & 47.93 & 52.98 & $<.001^{**}$ \\
      &\checkmark & 1.7B & \textbf{90.81} & 89.33          & $+1.48$ & 59.59 & 66.75 & $<.001^{**}$ \\
      & & 4B   & \underline{94.24} & 93.90          & $+0.34$ & 68.29 & 75.48 & $.105$                \\
      & & 8B   & \underline{95.62} & 95.51          & $+0.11$ & 74.81 & 80.15 & $.598$                \\
    \midrule

    \multirow{4}{*}{WinoGrande}
      &\checkmark & 0.6B & \textbf{70.80} & 67.80          & $+3.00$ & 56.82 & 57.61 & $.006^{**}$  \\
      &\checkmark & 1.7B & \underline{78.16} & 77.48          & $+0.68$ & 62.37 & 65.53 & $.518$                \\
      & & 4B   & \underline{87.21} & 87.15          & $+0.06$ & 66.67 & 67.74 & $1.000$               \\
      & & 8B   & \underline{89.53} & 89.25          & $+0.28$ & 69.89 & 72.21 & $.733$                \\
    \midrule

    \multirow{4}{*}{PIQA}
      &\checkmark & 0.6B & \textbf{77.80} & 74.70          & $+3.10$ & 68.34 & 71.87 & $.003^{**}$  \\
      &\checkmark & 1.7B & \textbf{85.04} & 82.59          & $+2.45$ & 74.86 & 77.48 & $.006^{**}$  \\
      & & 4B   & \underline{89.66} & 89.39          & $+0.27$ & 76.93 & 79.98 & $.754$                \\
      & & 8B   & 91.19          & \underline{92.11} & $-0.92$ & 79.92 & 82.05 & $.201$                \\
    \midrule\midrule

    \multirow{4}{*}{SciQ}
      &\checkmark & 0.6B & \textbf{90.70} & 88.50          & $+2.20$ & 75.60 & 86.00 & $.026^{*}$   \\
      &\checkmark & 1.7B & \underline{94.20} & 93.20          & $+1.00$ & 82.80 & 92.40 & $.203$                \\
      & & 4B   & 95.00          & \underline{95.10} & $-0.10$ & 90.80 & 93.00 & $1.000$               \\
      & & 8B   & 96.00          & \underline{96.90} & $-0.90$ & 92.20 & 95.80 & $.188$                \\
    \midrule
	
    \multirow{4}{*}{ARC-Challenge}
      &\checkmark & 0.6B & 63.61          & \underline{64.03} & $-0.42$ & 36.39 & 42.75 & $.817$                \\
      &\checkmark & 1.7B & 79.57          & \underline{79.66} & $-0.09$ & 43.09 & 55.71 & $1.000$               \\
      & & 4B   & 88.33          & \underline{89.70} & $-1.37$ & 56.31 & 64.03 & $.178$                \\
      & & 8B   & 91.07          & \underline{92.45} & $-1.38$ & 58.88 & 68.84 & $.121$                \\
    \bottomrule
  \end{tabular}
  \caption{
    Test accuracy on all five benchmarks. The upper block is the three primary commonsense
    benchmarks; the lower block is the two memorisation-intensive
    science benchmarks. \textsc{Cls-SFT} and \textsc{Label-Gen SFT} report
    native trained-format accuracy; Label-Gen LogProb and Gold-Only
    SFT report normalised log-probability accuracy on the
    candidate-sequence format (LP-Char for HellaSwag, PIQA, ARC-Challenge and SciQ, and 
    LP-Avg for WinoGrande). McNemar $p$-values compare paired
    correctness of \textsc{Cls-SFT} and Label-Gen SFT;
    significant entries are bolded ${}^{*}p<0.05$,
    ${}^{**}p<0.01$. Underlined entries mark the higher of
    \textsc{Cls-SFT} vs.\ Label-Gen SFT per row.
  }
  \label{tab:main_results_combined}
\end{table*}

\section{Results and Discussion}

\label{sec:results}

Table~\ref{tab:main_results_combined} reports test accuracies on all
five benchmarks across the four Qwen3 model sizes and three fine-tuning paradigms. Figure~\ref{fig:teaser} situates the commonsense results
against prior external baselines.
Detailed log-prob evaluations
(LP-Full, LP-Sum, LP-Avg, LP-Char) are reported in
Appendix~\ref{app:commonsense-results}. 

\subsection{Classification SFT is superior at tiny scale}
\label{sec:disc:tiny-scale}

The clearest pattern is that \textbf{Classification-head SFT is most
beneficial for TLMs}. At 0.6B, \textsc{Cls-SFT} improves over
Label-Gen SFT by $+2.45$, $+3.00$, $+3.10$, and $+2.20$ points on
HellaSwag, WinoGrande, PIQA, and SciQ; at 1.7B the gains remain
positive at $+1.48$, $+0.68$, $+2.45$, and $+1.00$. ARC-Challenge is
the only TLM-scale exception, and the difference is negligible
($-0.42$ at 0.6B, $-0.09$ at 1.7B, both non-significant with
$p\geq 0.8$). We attribute this to its small training split of only
$1{,}119$ examples. We interpret the broader pattern as evidence that
\textbf{tiny models benefit from being trained to compare candidate
sequences directly}. Under Label-Gen SFT the supervision is
concentrated on a single answer-letter token, whereas
\textsc{Cls-SFT} assigns an explicit scalar score to each candidate
sequence and optimises a softmax over those scores, directly exposing
the model to both gold and distractor candidates during training.

\subsection{Commonsense reasoning benchmarks}
\label{sec:disc:commonsense}

On HellaSwag, WinoGrande, and PIQA, \textsc{Cls-SFT} is the higher
of the two finetuning methods in every (dataset, model) row across all model scales, except PIQA-8B where Label-Gen SFT is better but not significant ($91.19$ vs.\ $92.11$, $\Delta = -0.92$, $p = .201$). At the TLM
scale the advantage is also statistically robust: five of the six TLM McNemar comparisons reach $p<0.01$ under the exact
binomial test, with WinoGrande-1.7B ($p=.518$) the only exception. 
\textbf{To the best of our knowledge, the \textsc{Cls-SFT} accuracies
reported here for Qwen3-0.6B and 1.7B are the highest
published for these models on HellaSwag, WinoGrande, and PIQA.}

\subsection{Memorization intensive benchmarks}
\label{sec:disc:science}

The science benchmarks split into two complementary cases. On
SciQ, which has a large training pool ($11.7$k examples)
and a clear knowledge-retrieval flavor, \textsc{Cls-SFT}
significantly outperforms Label-Gen SFT at 0.6B ($90.70\%$ vs.\
$88.50\%$, $p=.026$) and remains the higher-scoring method at 1.7B,
mirroring the TLM-scale pattern from the commonsense block. On
ARC-Challenge, Label-Gen SFT is
directionally higher at every scale, although none of the
differences is significant ($p\geq 0.12$ everywhere). We attribute
this to two compounding factors. First, ARC-Challenge has only
$1{,}119$ training examples, which limits how much either
fine-tuning objective can reshape behaviour. Second, the benchmark
is gated by retrieval of stored scientific facts rather than by
discrimination among plausible candidate texts. If the model does not possess the relevant fact or concept, changing the fine-tuning objective may not help as much. 

\subsection{Classification SFT at larger scale}
\label{sec:disc:large-scale}

The advantage of \textsc{Cls-SFT} shrinks sharply once we move past
the TLM regime. On the commonsense block, \textsc{Cls-SFT} is still
directionally ahead at 4B and 8B mostly, but
none of the 4B or 8B differences is statistically significant. The
single directional loss on commonsense is PIQA-8B. On the science
block, the pattern flips at scale: Label-Gen SFT is directionally
higher than \textsc{Cls-SFT} at 4B and 8B on both SciQ and
ARC-Challenge, but again every one of those differences is
non-significant. The narrowing therefore reflects \textbf{larger
models becoming better able to use the standard instruction-style
multiple-choice format}. Once the model has enough capacity to
reliably map the candidate texts in the prompt to the answer-label
token, the additional structure of an explicit candidate-scoring
objective stops mattering.

\subsection{Gold-Only continuation SFT}
\label{sec:disc:gold-only}

Across all five benchmarks the Gold-Only baseline trails both
\textsc{Cls-SFT} and Label-Gen SFT, often by ten points or more
(e.g., HellaSwag-1.7B: $66.75\%$ vs.\ $90.81\%$ for \textsc{Cls-SFT}).
Although Gold-Only SFT is the closest of the three paradigms to the
model's original pretraining objective (next-token prediction on the
gold context-candidate sequence, with the context masked from the
loss), it never directly supervises against distractors, leaving the
model with no signal for ranking wrong candidates below the right
one. For verifiable multi-choice tasks,
\textbf{supervising the contrast between gold and distractors is
more valuable than additional density on the gold candidate.}

\subsection{Fine-tuning objective determines the prediction interface}
\label{sec:disc:prediction-interface}

Our results also show that \textbf{fine-tuning changes which
prediction interface works best}. The same backbone can be used in
three ways (answer-letter generation through the LM head, candidate
log-probability scoring through the LM head, or sequence-level
scoring through a classification head), and after fine-tuning these
interfaces are no longer interchangeable.

When a \textsc{Cls-SFT}-trained model is evaluated through the
LM head instead of its native classification head, accuracy drops
sharply (e.g., Qwen3-1.7B+\textsc{Cls-SFT} on HellaSwag falls from
$90.81\%$ to $59.15\%$ under LP-Char; on PIQA, from $85.04\%$ to
$70.24\%$). The complementary swing appears for Label-Gen SFT:
re-evaluating the same adapter as a candidate scorer in the
\textsc{Cls-SFT}-style input format (Label-Gen LogProb) costs
roughly $25$ points on HellaSwag-4B ($93.90\%\to 68.29\%$) and
about $20$ points on WinoGrande-8B ($89.25\%\to 69.89\%$).

The fine-tuning objective and the evaluation interface are therefore
tightly coupled. Evaluating a fine-tuned
model through a different interface can substantially underestimate
the capability learned by that objective, and the swings are largest
for tiny models, where the choice of interface can shift reported
accuracy by tens of points.

\section{Conclusion}
\label{sec:conclusion}
We define Tiny Language Models, TLMs - as sub-3B language models motivated by practical
consumer-device memory constraints, and study how this regime should be
fine-tuned for verifiable multiple-choice reasoning. On a unified LoRA setup across Qwen3 0.6B--8B, our
classification-head formulation (\textsc{Cls-SFT}) beats
instruction-formatted label generation by $+2$ to $+3$ points on
every commonsense benchmark at 0.6B and 1.7B (five of six small-scale
comparisons significant at $p<0.01$), extends to SciQ at the TLM scale,
and is statistically neutral on ARC-Challenge. To our knowledge, the
resulting Qwen3-0.6B and Qwen3-1.7B numbers on HellaSwag, WinoGrande,
and PIQA are the highest published for these models.

For practitioners deploying TLMs on consumer hardware, our results
offer a concrete recipe: classification-head fine-tuning on a modest
LoRA budget leads to best performance. This makes the approach a safe default for verifiable
multi-choice tasks with models at TLM scale.

\section*{Limitations}
While we aim to provide a rigorous and controlled comparison of
fine-tuning interfaces for TLMs, our study necessarily makes several
scope and design choices. We acknowledge the following limitations.
\begin{itemize}\itemsep2pt

	\item \textbf{Scoring-head capacity.}
	      We use the standard \texttt{AutoModelForSequenceClassification}
	      head, which is a single linear projection from the final
	      non-padding hidden state to one scalar score. We do not add an
	      additional MLP or non-linear activation. This is appropriate for
	      cross-entropy training and is the simplest approach, which operates on raw logits, but a deeper
	      scoring head may provide additional capacity and could improve
	      accuracy. We leave this for future work.

	\item \textbf{Prompt-format sensitivity.}
	      We do not exhaustively explore prompt and answer-format choices.
	      For the Label-Gen SFT baseline, we use answer letters such
	      as A, B, C, and D. Other formats, including numbered options,
	      different chat templates, or dataset-specific instructions, may
	      affect the strength of the SFT baseline.

	\item \textbf{No few-shot inference.}
	      Our experiments focus on fine-tuned evaluation without few-shot
	      examples at inference time. Few-shot prompting may change the
	      relative performance of Label-Gen SFT and
	      candidate-sequence scoring.

	\item \textbf{No reasoning supervision.}
	      Recent small language models increasingly support reasoning-style
	      generation, but our experiments do not use chain-of-thought
	      supervision or reasoning traces. Our goal is to isolate the effect
	      of the candidate-scoring objective, but combining it with
	      reasoning-oriented training or inference may yield stronger results.

	\item \textbf{Limited model-family coverage.}
	      We focus on the Qwen3 model family to study scaling trends across a
	      consistent set of checkpoints. Due to compute constraints, we do not
	      repeat the full comparison across multiple model families.

	\item \textbf{Limited scale beyond 8B.}
	      We evaluate models up to the 7--8B parameter range. It remains
	      unclear whether the classification-head advantage persists for much
	      larger models, where Label-Gen SFT may already be strong
	      enough to exploit the multiple-choice prompt format.

	\item \textbf{Limited task and language coverage.}
	      Our evaluation focuses on English multiple-choice benchmarks, with
	      additional science-domain results reported separately. We do not
	      evaluate multilingual tasks, open-ended generation tasks, or
	      verifiable tasks where candidate sets must be generated rather than
	      provided by the dataset.

\end{itemize}
\section*{Ethical Considerations}
This study evaluates publicly available open-weight language models
on standardised, publicly available multi-choice benchmarks. No
private, sensitive, or human-subjects data were collected or used at
any stage of the work. All training and evaluation details, including
hyperparameters, prompt templates, and per-dataset configurations,
are reported in the main paper and appendices to support full
reproducibility.
\paragraph{AI Assistance:} We used Claude Code and ChatGPT for drafting LaTeX table
code, refining wording, and suggesting
code-organization patterns for the experimental pipeline. All generated content was reviewed and revised thoroughly by the authors to ensure accuracy and clarity.

\bibliography{custom}

\clearpage
\appendix

\section{Model and Device Memory}
\label{app:model-device-memory}

This appendix provides the hardware and memory context motivating our focus on small language models. Table~\ref{tab:model-memory-appendix} reports approximate model-weight memory requirements under common numerical formats, while Table~\ref{tab:consumer-device-memory} summarizes representative consumer-device memory capacities. These estimates are intended only as order-of-magnitude constraints: they exclude activations, optimizer states, framework overhead, and KV cache.
\begin{table}[H]
\centering
\scriptsize
\setlength{\tabcolsep}{5.2pt}
\renewcommand{\arraystretch}{1.12}
\resizebox{\columnwidth}{!}{%
\begin{tabular}{llrrrr}
\toprule
\textbf{Family} & \textbf{Model tier} & \textbf{Params} &
\textbf{FP32} & \textbf{BF16/FP16} & \textbf{FP8} \\
\midrule

\multicolumn{6}{l}{\textit{Qwen family}} \\
Qwen2.5 & 0.5B & 0.5B & 2.0GB  & 1.0GB  & 0.5GB \\
Qwen3   & 0.6B & 0.6B & 3.0GB  & 1.5GB  & 0.8GB \\
Qwen2.5 & 1.5B & 1.5B & 6.0GB  & 3.0GB  & 1.5GB \\
Qwen3   & 1.7B & 1.7B & 6.9GB  & 3.4GB  & 1.7GB \\
Qwen2.5 & 3B   & 3.0B & 12.0GB & 6.0GB  & 3.0GB \\
Qwen3   & 4B   & 4.0B & 16.1GB & 8.1GB  & 4.0GB \\
Qwen2.5 & 7B   & 7.0B & 28.0GB & 14.0GB & 7.0GB \\
Qwen3   & 8B   & 8.0B & 32.8GB & 16.4GB & 8.2GB \\

\midrule
\multicolumn{6}{l}{\textit{Llama family}} \\
Llama-3.2   & 1B & 1.0B & 5.0GB  & 2.5GB  & 1.2GB \\
Llama-3.2   & 3B & 3.0B & 12.8GB & 6.4GB  & 3.2GB \\
Llama-3/3.1 & 8B & 8.0B & 32.1GB & 16.1GB & 8.0GB \\

\midrule
\multicolumn{6}{l}{\textit{Gemma family}} \\
Gemma-3 & 1B & 1.0B & 4.0GB  & 2.0GB  & 1.0GB \\
Gemma   & 2B & 2.0B & 8.0GB  & 4.0GB  & 2.0GB \\
Gemma-2 & 2B & 2.0B & 8.0GB  & 4.0GB  & 2.0GB \\
Gemma-3 & 4B & 4.0B & 16.0GB & 8.0GB  & 4.0GB \\
Gemma-2 & 9B & 9.0B & 36.0GB & 18.0GB & 9.0GB \\

\midrule
\multicolumn{6}{l}{\textit{Phi family}} \\
Phi       & 1.5      & 1.3B & 5.2GB  & 2.6GB  & 1.3GB \\
Phi       & 2        & 2.7B & 10.8GB & 5.4GB  & 2.7GB \\
Phi       & 3.5-mini & 3.8B & 15.2GB & 7.6GB  & 3.8GB \\
Phi       & 4-mini   & 3.8B & 15.2GB & 7.6GB  & 3.8GB \\
Phi       & 3-small  & 7.0B & 28.0GB & 14.0GB & 7.0GB \\

\midrule
\multicolumn{6}{l}{\textit{Other open small-model families}} \\
MobileLLM & 0.6B & 0.6B & 2.4GB  & 1.2GB  & 0.6GB \\
MobileLLM & 1.5B & 1.5B & 6.0GB  & 3.0GB  & 1.5GB \\
OpenELM   & 1.1B & 1.1B & 4.4GB  & 2.2GB  & 1.1GB \\
OpenELM   & 3B   & 3.0B & 12.0GB & 6.0GB  & 3.0GB \\
SmolLM2   & 1.7B & 1.7B & 6.8GB  & 3.4GB  & 1.7GB \\
Mistral   & 7B   & 7.0B & 28.0GB & 14.0GB & 7.0GB \\

\bottomrule
\end{tabular}
}
\caption{
Approximate model-weight memory across common precision formats. For
the Qwen3 and Llama rows used in our main experiments, values match the
Hugging Face safetensor shard sizes reported in
Table~\ref{tab:model-memory}. Other rows are included as approximate
family-level references. All values exclude activations, optimizer
states, runtime overhead, and KV cache, so actual inference or training
memory is higher.
}
\label{tab:model-memory-appendix}
\end{table}

\begin{table}[H]
\centering
\footnotesize
\setlength{\tabcolsep}{4.0pt}
\renewcommand{\arraystretch}{1.12}
\resizebox{\columnwidth}{!}{%
\begin{tabularx}{\textwidth}{lllX}
\toprule
\textbf{Tier} & \textbf{Device / configuration} &
\textbf{Memory} & \textbf{Type} \\
\midrule
\multicolumn{4}{l}{\textit{8GB and below: common entry or mid-range COTS memory limits}} \\
\midrule
Phone & iPhone 17 base & 8GB & Unified system memory \\
Phone & Google Pixel 8a & 8GB & Unified system memory \\
Phone & Samsung Galaxy A56, lower configuration & 8GB & Unified system memory \\
Phone & Redmi Note 14 Pro, lower configuration & 8GB & Unified system memory \\
Phone & Nothing Phone (2a), lower configuration & 8GB & Unified system memory \\
Laptop & HP Victus 15 with RTX 3050-class GPU & 4GB VRAM & Dedicated GPU memory \\
Laptop & NVIDIA RTX 4050 laptop GPU systems & 6GB VRAM & Dedicated GPU memory \\
Laptop & ASUS ROG Zephyrus G14 with RTX 4060 & 8GB VRAM & Dedicated GPU memory \\
Laptop & HP OMEN / Victus with RTX 4070 & 8GB VRAM & Dedicated GPU memory \\
Laptop & Dell XPS 16 with RTX 5070 & 8GB VRAM & Dedicated GPU memory \\
Laptop & NVIDIA RTX 5060 / 5070 laptop GPU systems & 8GB VRAM & Dedicated GPU memory \\
\midrule
\multicolumn{4}{l}{\textit{Above 8GB: higher-end phones, UMA laptops, and gaming/creator laptops}} \\
\midrule
Phone & iPhone 17 Air / Pro / Pro Max & 12GB & Unified system memory \\
Phone & Google Pixel 9 & 12GB & Unified system memory \\
Phone & Google Pixel 9 Pro / Pro XL & 16GB & Unified system memory \\
Phone & Samsung Galaxy S25 Ultra & 12GB & Unified system memory \\
Phone & Samsung Galaxy A56, higher configuration & 12GB & Unified system memory \\
Phone & OnePlus 13 & 12GB / 16GB & Unified system memory \\
Phone & OnePlus Nord 4 & 12GB / 16GB & Unified system memory \\
Phone & Xiaomi 15 Ultra & 16GB & Unified system memory \\
Phone & OPPO Find X8 Pro & 16GB & Unified system memory \\
Phone & Nothing Phone (3) & 12GB / 16GB & Unified system memory \\
Phone & Nothing Phone (2a), higher configuration & 12GB & Unified system memory \\
\midrule
Laptop & MacBook Air, Apple Silicon & 16GB / 24GB / 32GB & Unified memory \\
Laptop & MacBook Pro, Apple Silicon Pro/Max & 24GB--128GB & Unified memory \\
Laptop & Microsoft Surface Laptop 13.8'' / 15'' & 16GB / 32GB / 64GB & System memory \\
Laptop & Dell XPS 13 & 16GB / 32GB & System memory \\
Laptop & Dell XPS 16 & 16GB / 32GB / 64GB & System memory \\
Laptop & HP OmniBook Ultra / AI PC class & 16GB / 32GB & System memory \\
Laptop & HP Spectre / premium convertible class & 16GB / 32GB & System memory \\
Laptop & ASUS ROG Zephyrus / similar creator laptops & 16GB / 32GB & System memory plus optional dGPU VRAM \\
\midrule
Laptop & NVIDIA RTX 4080 laptop GPU systems & 12GB VRAM & Dedicated GPU memory \\
Laptop & NVIDIA RTX 4090 laptop GPU systems & 16GB VRAM & Dedicated GPU memory \\
Laptop & NVIDIA RTX 5070 Ti laptop GPU systems & 12GB VRAM & Dedicated GPU memory \\
Laptop & NVIDIA RTX 5080 laptop GPU systems & 16GB VRAM & Dedicated GPU memory \\
Laptop & NVIDIA RTX 5090 laptop GPU systems & 24GB VRAM & Dedicated GPU memory \\
\bottomrule
\end{tabularx}
}
\caption{
Representative memory capacities of recent consumer devices.
Phone and Apple-silicon laptop memory is unified and shared with the
operating system, applications, model weights, activations, and KV
cache. NVIDIA laptop GPUs use dedicated VRAM. \textbf{For laptops with
discrete GPUs, system RAM is not a substitute for GPU VRAM for practical
local LLM inference: the model must fit primarily in GPU memory to run
at usable speed.} Apple-silicon unified memory is a partial exception
because the CPU and GPU share the same memory pool, but it is still
shared with the rest of the system. These constraints motivate the
sub-3B TLM regime: larger 4B--8B models often require high-end
memory configurations, aggressive quantization, or offloading.
}
\label{tab:consumer-device-memory}
\end{table}

\clearpage
\onecolumn
\begin{multicols}{2}
\section{Input Formats and Training Templates}
\label{app:input-formats}

This appendix gives concrete examples and abstract templates for the three training paradigms. The examples illustrate how the same supervised item is converted into candidate-sequence scoring, answer-label generation, and gold-only continuation supervised fine-tuning. We include these details to make the comparison reproducible and to clarify exactly where the training loss is applied in each paradigm.

\subsection{Concrete Input Examples}
\label{app:example-input-formats}

Table~\ref{tab:appendix_example_formats} shows one representative example from each primary commonsense benchmark. In \textsc{Cls}, every answer option is converted into a separate candidate sequence and scored with the classification head. In Label-Gen SFT, all choices are placed in a chat-template prompt and the model is supervised to generate the answer label. In Gold-Only SFT, only the gold candidate sequence is used for causal language-model fine-tuning.
\end{multicols}

\begin{table}[h]
\centering
\scriptsize
\setlength{\tabcolsep}{5pt}
\renewcommand{\arraystretch}{1.18}
\arrayrulecolor{rulegray}
\begin{tabularx}{\textwidth}{|D|C|S|T|}
\hline
\rowcolor{headerbg}
\textbf{Dataset} &
\textbf{\textsc{Cls}: candidate-sequence scoring} &
\textbf{Label-Gen SFT: answer-label generation} &
\textbf{Gold-Only SFT: gold-only continuation} \\
\hline

HellaSwag &
\textit{All candidates are scored independently:}\par\smallskip
A. \ctx{Roof shingle removal: A man is sitting on a roof. He} \cand{is using wrap to wrap a pair of skis.}\par
B. \ctx{Roof shingle removal: A man is sitting on a roof. He} \cand{is ripping level tiles off.}\par
C. \ctx{Roof shingle removal: A man is sitting on a roof. He} \cand{is holding a Rubik's cube.}\par
D. \ctx{Roof shingle removal: A man is sitting on a roof. He} \gold{starts pulling up roofing on a roof.}\par\smallskip
\textit{Training: cross-entropy over four scalar classification-head scores.}
&
\texttt{<|im\_start|>user}\par
Choose the most plausible ending.\par\smallskip
Context: \ctx{Roof shingle removal: A man is sitting on a roof. He}\par\smallskip
A. is using wrap to wrap a pair of skis.\par
B. is ripping level tiles off.\par
C. is holding a Rubik's cube.\par
D. starts pulling up roofing on a roof.\par\smallskip
Answer with only one letter: A, B, C, or D.\par
Answer:\texttt{<|im\_end|>}\par
\texttt{<|im\_start|>assistant} \target{D}\par\smallskip
\textit{Training: next-token prediction on the answer label.}
&
\textit{Only the gold continuation is used:}\par\smallskip
\ctx{Roof shingle removal: A man is sitting on a roof. He} \gold{starts pulling up roofing on a roof.}\par\smallskip
\textit{Training: next-token prediction only on the gold ending tokens; context tokens are masked from the loss.}
\\
\hline

WinoGrande &
\textit{Each option is substituted into the blank:}\par\smallskip
A. \ctx{Sarah was a much better surgeon than Maria so} \cand{Sarah} \ctx{always got the easier cases.}\par
B. \ctx{Sarah was a much better surgeon than Maria so} \gold{Maria} \ctx{always got the easier cases.}\par\smallskip
\textit{Training: cross-entropy over two scalar classification-head scores.}
&
\texttt{<|im\_start|>user}\par
Fill in the blank with the most appropriate option.\par\smallskip
Sentence: \ctx{Sarah was a much better surgeon than Maria so \_ always got the easier cases.}\par\smallskip
A. Sarah\par
B. Maria\par\smallskip
Answer with only one letter: A or B.\par
Answer:\texttt{<|im\_end|>}\par
\texttt{<|im\_start|>assistant} \target{B}\par\smallskip
\textit{Training: next-token prediction on the answer label.}
&
\textit{Only the gold substituted sentence is used:}\par\smallskip
\ctx{Sarah was a much better surgeon than Maria so} \gold{Maria} \ctx{always got the easier cases.}\par\smallskip
\textit{Training: next-token prediction on the substituted option and remaining suffix; prefix tokens are masked.}
\\
\hline

PIQA &
\textit{Each solution is paired with the same goal:}\par\smallskip
A. \ctx{Question: How do I ready a guinea pig cage for its new occupants? Answer:}
\gold{Provide the guinea pig with a cage full of a few inches of bedding made of ripped paper strips, you will also need to supply it with a water bottle and a food dish.}\par\smallskip
B. \ctx{Question: How do I ready a guinea pig cage for its new occupants? Answer:}
\cand{Provide the guinea pig with a cage full of a few inches of bedding made of ripped jeans material, you will also need to supply it with a water bottle and a food dish.}\par\smallskip
\textit{Training: cross-entropy over two scalar classification-head scores.}
&
\texttt{<|im\_start|>user}\par
Choose the most appropriate solution for the given goal.\par\smallskip
Goal: \ctx{How do I ready a guinea pig cage for its new occupants?}\par\smallskip
A. Provide the guinea pig with a cage full of a few inches of bedding made of ripped paper strips...\par
B. Provide the guinea pig with a cage full of a few inches of bedding made of ripped jeans material...\par\smallskip
Answer with only one letter: A or B.\par
Answer:\texttt{<|im\_end|>}\par
\texttt{<|im\_start|>assistant} \target{A}\par\smallskip
\textit{Training: next-token prediction on the answer label.}
&
\textit{Only the gold solution is used:}\par\smallskip
\ctx{Question: How do I ready a guinea pig cage for its new occupants? Answer:}
\gold{Provide the guinea pig with a cage full of a few inches of bedding made of ripped paper strips, you will also need to supply it with a water bottle and a food dish.}\par\smallskip
\textit{Training: next-token prediction only on the gold solution tokens; question/prefix tokens are masked.}
\\
\hline

\end{tabularx}
\arrayrulecolor{black}
\captionof{table}{
Concrete examples of the three training paradigms. Blue text denotes shared context or question text, orange denotes incorrect candidate content, green denotes the gold candidate, and purple denotes the supervised answer label. \textsc{Cls} scores all candidates with a scalar classification head. Label-Gen SFT trains the model to emit a label token. Gold-Only SFT uses only the gold candidate sequence and applies next-token prediction only to the candidate/ending tokens.
}
\label{tab:appendix_example_formats}
\end{table}

\clearpage
\onecolumn
\begin{multicols}{2}
\subsection{Abstract Instruction Templates}
\label{app:instruction-templates}

Table~\ref{tab:appendix_instruction_templates} abstracts away from individual examples and shows the dataset-specific templates used to construct inputs. These templates are the actual formatting rules used to convert each supervised example into the three training objectives.
\end{multicols}

\begin{table}[H]
\centering
\scriptsize
\setlength{\tabcolsep}{5pt}
\renewcommand{\arraystretch}{1.18}
\arrayrulecolor{rulegray}
\begin{tabularx}{\textwidth}{|D|C|S|T|}
\hline
\rowcolor{headerbg}
\textbf{Dataset} &
\textbf{\textsc{Cls}: candidate-sequence format} &
\textbf{Label-Gen SFT: chat-template format} &
\textbf{Gold-Only SFT: gold-only continuation format} \\
\hline

HellaSwag &
\textit{Four candidates are constructed:}\par\smallskip
A. \ctx{\{ctx\}} \cand{\{ending\_0\}}\par
B. \ctx{\{ctx\}} \cand{\{ending\_1\}}\par
C. \ctx{\{ctx\}} \cand{\{ending\_2\}}\par
D. \ctx{\{ctx\}} \cand{\{ending\_3\}}\par\smallskip
\textit{Each candidate receives one scalar classification-head score; cross-entropy is applied over the four scores.}
&
\texttt{<|im\_start|>user}\par
Choose the most plausible ending for the given context.\par\smallskip
Context: \ctx{\{ctx\}}\par\smallskip
A. \{ending\_0\}\par
B. \{ending\_1\}\par
C. \{ending\_2\}\par
D. \{ending\_3\}\par\smallskip
Answer with only one letter: A, B, C, or D.\par
Answer:\texttt{<|im\_end|>}\par
\texttt{<|im\_start|>assistant} \target{\{A|B|C|D\}}\par\smallskip
\textit{Loss is applied only to the answer-label token.}
&
\ctx{\{ctx\}} \gold{\{gold ending\}}\par\smallskip
\textit{Only the correct ending is used. The context is masked, and next-token prediction loss is applied over the gold ending tokens.}
\\
\hline

WinoGrande &
\textit{The sentence is split at the blank:}\par\smallskip
A. \ctx{\{prefix\}} \cand{\{option\_1\}} \ctx{\{suffix\}}\par
B. \ctx{\{prefix\}} \cand{\{option\_2\}} \ctx{\{suffix\}}\par\smallskip
\textit{Each substituted sentence receives one scalar classification-head score; cross-entropy is applied over the two scores.}
&
\texttt{<|im\_start|>user}\par
Fill in the blank in the following sentence with the most appropriate option.\par\smallskip
Sentence: \ctx{\{sentence with \texttt{\_} blank\}}\par\smallskip
A. \{option\_1\}\par
B. \{option\_2\}\par\smallskip
Answer with only one letter: A or B.\par
Answer:\texttt{<|im\_end|>}\par
\texttt{<|im\_start|>assistant} \target{\{A|B\}}\par\smallskip
\textit{Loss is applied only to the answer-label token.}
&
\ctx{\{prefix\}} \gold{\{gold option\}} \ctx{\{suffix\}}\par\smallskip
\textit{Only the correct substituted sentence is used. The prefix before the blank is masked, and next-token prediction loss is applied to the gold option and remaining suffix tokens.}
\\
\hline

PIQA &
\textit{Two question-answer candidates are constructed:}\par\smallskip
A. \ctx{Question: \{goal\}\textbackslash nAnswer:} \cand{\{sol\_1\}}\par
B. \ctx{Question: \{goal\}\textbackslash nAnswer:} \cand{\{sol\_2\}}\par\smallskip
\textit{Each solution receives one scalar classification-head score; cross-entropy is applied over the two scores.}
&
\texttt{<|im\_start|>user}\par
Choose the most appropriate solution for the given goal.\par\smallskip
Goal: \ctx{\{goal\}}\par\smallskip
A. \{sol\_1\}\par
B. \{sol\_2\}\par\smallskip
Answer with only one letter: A or B.\par
Answer:\texttt{<|im\_end|>}\par
\texttt{<|im\_start|>assistant} \target{\{A|B\}}\par\smallskip
\textit{Loss is applied only to the answer-label token.}
&
\ctx{Question: \{goal\}\textbackslash nAnswer:} \gold{\{gold solution\}}\par\smallskip
\textit{Only the correct solution is used. The question and \texttt{Answer:} prefix are masked, and next-token prediction loss is applied over the gold solution tokens.}
\\
\hline

\end{tabularx}
\arrayrulecolor{black}
\caption{Abstract input templates used by the three paradigms. \textsc{Cls} converts every answer option into a complete candidate sequence and trains a scalar scoring head with cross-entropy over candidates. Label-Gen SFT presents all choices in a chat-template prompt and supervises only the answer label. Gold-Only SFT uses the same candidate construction as \textsc{Cls}, but keeps only the gold candidate and trains the causal LM with next-token prediction over the gold answer/ending tokens.}
\label{tab:appendix_instruction_templates}
\end{table}

\clearpage
\twocolumn
\section{Detailed Evaluation Results}
\subsection{LoRA and Training Hyperparameters}
\label{app:lora_hyperparams}

All fine-tuning runs use Low-Rank Adaptation (LoRA) over all linear
layers of the transformer backbone. We use rank $r=16$, scaling factor
$\alpha=32$, and no bias adaptation. For the causal language-modeling
baselines, \textsc{LabelGen} and \textsc{GoldOnly}, we use LoRA dropout
$0.05$. For \textsc{Cls}, we use the sequence-classification LoRA setup
with dropout $0.0$ (as we want each candidate to go through the same LoRA updates) and save the scalar classification score head using
\texttt{modules\_to\_save = ["score"]}. Thus, the newly initialized
classification head is trained in full together with the LoRA adapters,
while the base model parameters remain frozen.

\begin{table}[ht]
\centering
\scriptsize
\setlength{\tabcolsep}{5pt}
\renewcommand{\arraystretch}{1.12}
\begin{tabular}{lccc}
\toprule
\textbf{Setting} & \textbf{\textsc{Cls}} & \textbf{\textsc{LabelGen}} & \textbf{\textsc{GoldOnly}} \\
\midrule
Backbone model & Qwen3 & Qwen3 & Qwen3 \\
HF model class & Seq. classification & Causal LM & Causal LM \\
LoRA task type & \texttt{SEQ\_CLS} & \texttt{CAUSAL\_LM} & \texttt{CAUSAL\_LM} \\
Target modules & all linear & all linear & all linear \\
LoRA rank $r$ & 16 & 16 & 16 \\
LoRA $\alpha$ & 32 & 32 & 32 \\
LoRA dropout & 0.0 & 0.05 & 0.05 \\
Bias adaptation & none & none & none \\
Saved extra modules & score head & -- & -- \\
\bottomrule
\end{tabular}
\caption{
LoRA configuration used for the three fine-tuning paradigms. The
\textsc{Cls} setting trains the scalar sequence-classification score
head in addition to LoRA adapters. \textsc{LabelGen} and
\textsc{GoldOnly} keep the causal language-modeling head and train only
LoRA adapters.
}
\label{tab:lora_hyperparams}
\end{table}

Unless otherwise stated, all runs use the same optimizer and schedule:
2 epochs, learning rate $2\times10^{-4}$, per-device batch size 8,
gradient accumulation over 8 steps, effective batch size 64, cosine
learning-rate decay, warmup ratio $0.1$ for \textsc{Cls} and $0.03$ for
the causal-LM baselines, weight decay $0.01$, and gradient clipping at
maximum norm $1.0$.

\subsection{Evaluation Protocol}
\label{app:evaluation_protocols}
Each trained model is evaluated in its native format. For
\textsc{Cls}, we report the accuracy of the classification-head argmax
over candidate scores. For Label-Gen SFT, we report next-token answer
accuracy over the answer-letter tokens. For Gold-Only SFT, we evaluate
candidate sequences using language-model log-probability scoring.

We additionally evaluate the Label-Gen SFT adapter as a candidate scorer
by applying language-model log-probability scoring to the same
candidate sequences used by \textsc{Cls}; we denote this evaluation as
Label-Gen LogProb. This is not a separately trained model, but an
apples-to-apples test of whether an instruction-tuned adapter also
improves candidate ranking.

We report the following log-probability-based candidate-scoring metrics:

\begin{itemize}\itemsep0pt
	\item \textbf{LP-Full:} log-probability of the full candidate
	      sequence.
	\item \textbf{LP-Sum:} sum of suffix token log-probabilities
	      conditioned on the context.
	\item \textbf{LP-Avg:} suffix log-probability normalized by suffix
	      token count.
	\item \textbf{LP-Char:} suffix log-probability normalized by suffix
	      character length.
\end{itemize}
\label{app:detailed-results}

This appendix reports the full set of evaluation metrics used in the paper. The main text focuses on the native accuracy of the trained objective, while the tables below also include language-model log-probability based re-scoring variants. These additional columns help separate improvements in answer-label generation from improvements in candidate-sequence ranking.

\subsection{Commonsense Benchmarks}
\label{app:commonsense-results}

Table~\ref{tab:main_results} gives the full results on the three primary commonsense benchmarks. For each dataset and model size, we compare the proposed classification-head objective against instruction-style supervised fine-tuning and continuation language-model fine-tuning.


\clearpage
\onecolumn
\begin{table}[H]
	\centering
	\scriptsize
	\setlength{\tabcolsep}{4pt}
	\renewcommand{\arraystretch}{1.15}
	\begin{tabular}{|l|l|c|c|c|c|c|c|c|}
		\hline
		\textbf{Dataset} & \textbf{Model}              & \textbf{Stage}  & \textbf{Trained} & \textbf{LP-Full} & \textbf{LP-Sum} & \textbf{LP-Avg} & \textbf{LP-Char}                           & \textbf{McNemar $p$\,$\dagger$} \\
		\hline\hline

		\multirow{16}{*}{HellaSwag}
		                 & \multirow{4}{*}{Qwen3-0.6B}
		                 & \textsc{Cls}                & \textbf{80.95}  & 36.39            & 36.36            & 47.76           & 48.19           & \multirow{4}{*}{$2.8\!\times\!10^{-11**}$}                                   \\ \cline{3-8}
		                 &                             & Label-Gen SFT & 78.50            & 60.07            & 64.10           & 57.52           & 56.29                                      &                                 \\ \cline{3-8}
		                 &                             & Label-Gen LogProb & ---              & 36.54            & 36.37           & 45.97           & 47.93                                      &                                 \\ \cline{3-8}
		                 &                             & Gold-Only SFT & ---              & 40.75            & 40.90           & 50.99           & 52.98                                      &                                 \\ \cline{2-9}
		                 & \multirow{4}{*}{Qwen3-1.7B}
		                 & \textsc{Cls}                & \textbf{90.81}  & 43.40            & 43.52            & 57.86           & 59.15           & \multirow{4}{*}{$8.5\!\times\!10^{-8**}$}                                    \\ \cline{3-8}
		                 &                             & Label-Gen SFT & 89.33            & 80.48            & 82.29           & 80.94           & 80.42                                      &                                 \\ \cline{3-8}
		                 &                             & Label-Gen LogProb & ---              & 45.01            & 45.14           & 57.81           & 59.59                                      &                                 \\ \cline{3-8}
		                 &                             & Gold-Only SFT & ---              & 50.09            & 50.10           & 64.71           & 66.75                                      &                                 \\ \cline{2-9}
		                 & \multirow{4}{*}{Qwen3-4B}
		                 & \textsc{Cls}                & \textbf{94.24}  & 46.23            & 46.32            & 63.94           & 65.46           & \multirow{4}{*}{$0.105$ n.s.}                                                \\ \cline{3-8}
		                 &                             & Label-Gen SFT & 93.90            & 91.10            & 92.94           & 91.73           & 91.29                                      &                                 \\ \cline{3-8}
		                 &                             & Label-Gen LogProb & ---              & 50.76            & 50.75           & 65.57           & 68.29                                      &                                 \\ \cline{3-8}
		                 &                             & Gold-Only SFT & ---              & 57.50            & 57.49           & 73.32           & 75.48                                      &                                 \\ \cline{2-9}
		                 & \multirow{4}{*}{Qwen3-8B}
		                 & \textsc{Cls}                & \textbf{95.62}  & 54.12            & 54.12            & 73.72           & 75.23           & \multirow{4}{*}{$0.598$ n.s. }                                               \\ \cline{3-8}
		                 &                             & Label-Gen SFT & 95.51            & 95.25            & 95.25           & 90.27           & 88.36                                      &                                 \\ \cline{3-8}
		                 &                             & Label-Gen LogProb & ---              & 55.51            & 55.51           & 72.37           & 74.81                                      &                                 \\ \cline{3-8}
		                 &                             & Gold-Only SFT & ---              & 61.65            & 61.65           & 78.08           & 80.15                                      &                                 \\
		\hline\hline

		\multirow{16}{*}{WinoGrande}
		                 & \multirow{4}{*}{Qwen3-0.6B}
		                 & \textsc{Cls}                & \textbf{70.80}  & 62.37            & 61.80            & 61.40           & ---             & \multirow{4}{*}{$0.0058^{**}$}                                               \\ \cline{3-8}
		                 &                             & Label-Gen SFT & 67.80            & 67.12            & 65.65           & 65.48           & ---                                        &                                 \\ \cline{3-8}
		                 &                             & Label-Gen LogProb & ---              & 57.56            & 57.50           & 56.82           & ---                                        &                                 \\ \cline{3-8}
		                 &                             & Gold-Only SFT & ---              & 57.61            & 58.23           & 57.61           & ---                                        &                                 \\ \cline{2-9}
		                 & \multirow{4}{*}{Qwen3-1.7B}
		                 & \textsc{Cls}                & \textbf{78.16}  & 67.01            & 67.01            & 66.16           & ---             & \multirow{4}{*}{$0.518$ n.s.}                                                \\ \cline{3-8}
		                 &                             & Label-Gen SFT & 77.48            & 76.51            & 76.00           & 75.95           & ---                                        &                                 \\ \cline{3-8}
		                 &                             & Label-Gen LogProb & ---              & 63.38            & 62.93           & 62.37           & ---                                        &                                 \\ \cline{3-8}
		                 &                             & Gold-Only SFT & ---              & 66.33            & 66.33           & 65.53           & ---                                        &                                 \\ \cline{2-9}
		                 & \multirow{4}{*}{Qwen3-4B}
		                 & \textsc{Cls}                & \textbf{87.21}  & 73.34            & 73.34            & 72.33           & ---             & \multirow{4}{*}{$1.000$ n.s.}                                                \\ \cline{3-8}
		                 &                             & Label-Gen SFT & 87.15            & 84.10            & 83.98           & 84.38           & ---                                        &                                 \\ \cline{3-8}
		                 &                             & Label-Gen LogProb & ---              & 67.23            & 67.52           & 66.67           & ---                                        &                                 \\ \cline{3-8}
		                 &                             & Gold-Only SFT & ---              & 68.42            & 68.59           & 67.74           & ---                                        &                                 \\ \cline{2-9}
		                 & \multirow{4}{*}{Qwen3-8B}
		                 & \textsc{Cls}                & \textbf{89.53}  & 73.85            & 73.63            & 72.50           & ---             & \multirow{4}{*}{$0.733$ n.s.}                                                \\ \cline{3-8}
		                 &                             & Label-Gen SFT & 89.25            & 84.15            & 80.59           & 80.25           & ---                                        &                                 \\ \cline{3-8}
		                 &                             & Label-Gen LogProb & ---              & 69.84            & 69.84           & 69.89           & ---                                        &                                 \\ \cline{3-8}
		                 &                             & Gold-Only SFT & ---              & 71.99            & 72.78           & 72.21           & ---                                        &                                 \\
		\hline\hline

		\multirow{16}{*}{PIQA}
		                 & \multirow{4}{*}{Qwen3-0.6B}
		                 & \textsc{Cls}                & \textbf{77.80}  & 65.56            & 65.23            & 65.56           & 66.27           & \multirow{4}{*}{$0.0034^{**}$}                                               \\ \cline{3-8}
		                 &                             & Label-Gen SFT & 74.70            & 73.83            & 73.01           & 72.63           & 72.63                                      &                                 \\ \cline{3-8}
		                 &                             & Label-Gen LogProb & ---              & 67.41            & 67.41           & 67.19           & 68.34                                      &                                 \\ \cline{3-8}
		                 &                             & Gold-Only SFT & ---              & 72.03            & 72.03           & 71.33           & 71.87                                      &                                 \\ \cline{2-9}
		                 & \multirow{4}{*}{Qwen3-1.7B}
		                 & \textsc{Cls}                & \textbf{85.04}  & 69.26            & 69.37            & 69.75           & 70.24           & \multirow{4}{*}{$0.0060^{**}$}                                               \\ \cline{3-8}
		                 &                             & Label-Gen SFT & 82.59            & 59.14            & 57.89           & 58.54           & 58.16                                      &                                 \\ \cline{3-8}
		                 &                             & Label-Gen LogProb & ---              & 73.07            & 73.45           & 74.27           & 74.86                                      &                                 \\ \cline{3-8}
		                 &                             & Gold-Only SFT & ---              & 76.93            & 76.82           & 77.26           & 77.48                                      &                                 \\ \cline{2-9}
		                 & \multirow{4}{*}{Qwen3-4B}
		                 & \textsc{Cls}                & \textbf{89.66}  & 74.65            & 74.86            & 75.30           & 76.77           & \multirow{4}{*}{$0.754$ n.s.}                                                \\ \cline{3-8}
		                 &                             & Label-Gen SFT & 89.39            & 88.30            & 88.90           & 89.12           & 88.79                                      &                                 \\ \cline{3-8}
		                 &                             & Label-Gen LogProb & ---              & 76.06            & 75.79           & 76.33           & 76.93                                      &                                 \\ \cline{3-8}
		                 &                             & Gold-Only SFT & ---              & 79.22            & 78.99           & 80.09           & 79.98                                      &                                 \\ \cline{2-9}
		                 & \multirow{4}{*}{Qwen3-8B}
		                 & \textsc{Cls}                & 91.19           & 75.79            & 75.95            & 76.88           & 77.15           & \multirow{4}{*}{$0.201$ n.s.}                                                \\ \cline{3-8}
		                 &                             & Label-Gen SFT & \textbf{92.11}   & 91.13            & 90.37           & 88.47           & 87.87                                      &                                 \\ \cline{3-8}
		                 &                             & Label-Gen LogProb & ---              & 78.45            & 78.40           & 79.82           & 79.92                                      &                                 \\ \cline{3-8}
		                 &                             & Gold-Only SFT & ---              & 81.56            & 81.45           & 82.32           & 82.05                                      &                                 \\
		\hline
	\end{tabular}
	\caption{Test-set accuracy (\%) for the three commonsense benchmarks
		under all four evaluated stages. \textsc{Trained} reports \textsc{Cls}
		head accuracy for Stage 1 rows and \textsc{NextTok} accuracy for
		Stage 2 rows. Label-Gen LogProb re-scores the Stage 2 SFT adapter using
		the \textsc{Cls} log-prob format for an apples-to-apples comparison.
		LP variants are defined in Appendix~\ref{app:evaluation_protocols}. \textbf{Bold} marks
		the highest accuracy per (dataset, model) block. $\dagger$~McNemar
		exact two-sided binomial $p$-value comparing \textsc{Cls} and
		Label-Gen SFT on per-example correctness; ${}^{**}p<0.01$,
		${}^{*}p<0.05$. All rows use the unified setup of Appendix~\ref{app:lora_hyperparams}.}
	\label{tab:main_results}
\end{table}
\clearpage
\twocolumn
\FloatBarrier

\subsection{Log-Probability Normalization Effects}
\label{app:length-normalization}

The log-probability evaluations reveal how sensitive candidate ranking is to sequence-length normalization. Across HellaSwag and PIQA, LP-Char is often stronger than LP-Avg for the candidate-sequence evaluations most relevant to our analysis, namely \textsc{Cls} and Label-Gen LogProb. This suggests that character-level normalization can better control for answer-length effects in these datasets. The pattern is not universal: native Label-Gen SFT log-probability scores often prefer LP-Avg, and WinoGrande does not use LP-Char in our evaluation.

\paragraph{SFT does not necessarily transfer to candidate scoring.}
Label-Gen LogProb is consistently lower than native Label-Gen SFT accuracy, even though both use the same trained adapter. This indicates that Label-Gen SFT primarily improves the model's ability to emit the correct answer label in the chat-template format, but does not reliably improve its ability to rank the candidate answer sequences themselves. Gold-Only SFT is usually stronger than Label-Gen LogProb under normalized log-probability scoring, but remains far below \textsc{Cls}. This suggests that increasing the likelihood of the gold continuation alone helps candidate scoring somewhat, but is weaker than directly training a discriminative objective over correct and incorrect candidates.

\subsection{Science Benchmarks}
\label{app:science-results}

Table~\ref{tab:science_results} reports the same evaluation suite on ARC-Challenge and SciQ. These benchmarks are used as additional science-domain evaluations rather than the primary benchmark suite. 

\clearpage
\onecolumn
\begin{table}[H]
	\centering
	\scriptsize
	\setlength{\tabcolsep}{4pt}
	\renewcommand{\arraystretch}{1.15}
	\begin{tabular}{|l|l|c|c|c|c|c|c|c|}
		\hline
		\textbf{Dataset} & \textbf{Model}              & \textbf{Stage}  & \textbf{Trained} & \textbf{LP-Full} & \textbf{LP-Sum} & \textbf{LP-Avg} & \textbf{LP-Char}              & \textbf{McNemar $p$\,$\dagger$} \\
		\hline\hline

		\multirow{16}{*}{ARC-Challenge}
		                 & \multirow{4}{*}{Qwen3-0.6B}
		                 & \textsc{Cls}                & 63.61           & 36.57            & 36.48            & 40.17           & 39.31           & \multirow{4}{*}{$0.817$ n.s.}                                   \\ \cline{3-8}
		                 &                             & Label-Gen SFT & \textbf{64.03}   & 63.00            & 63.52           & 63.78           & 64.03                         &                                 \\ \cline{3-8}
		                 &                             & Label-Gen LogProb & ---              & 34.25            & 34.68           & 37.17           & 36.39                         &                                 \\ \cline{3-8}
		                 &                             & Gold-Only SFT & ---              & 39.23            & 39.74           & 44.21           & 42.75                         &                                 \\ \cline{2-9}
		                 & \multirow{4}{*}{Qwen3-1.7B}
		                 & \textsc{Cls}                & 79.57           & 52.70            & 52.70            & 55.11           & 56.31           & \multirow{4}{*}{$1.000$ n.s.}                                   \\ \cline{3-8}
		                 &                             & Label-Gen SFT & \textbf{79.66}   & 78.80            & 79.48           & 79.40           & 79.57                         &                                 \\ \cline{3-8}
		                 &                             & Label-Gen LogProb & ---              & 39.66            & 39.83           & 42.40           & 43.09                         &                                 \\ \cline{3-8}
		                 &                             & Gold-Only SFT & ---              & 51.59            & 51.33           & 56.48           & 55.71                         &                                 \\ \cline{2-9}
		                 & \multirow{4}{*}{Qwen3-4B}
		                 & \textsc{Cls}                & 88.33           & 53.65            & 53.73            & 56.14           & 58.11           & \multirow{4}{*}{$0.178$ n.s.}                                   \\ \cline{3-8}
		                 &                             & Label-Gen SFT & \textbf{89.70}   & 89.53            & 89.79           & 89.79           & 89.70                         &                                 \\ \cline{3-8}
		                 &                             & Label-Gen LogProb & ---              & 52.10            & 52.53           & 54.68           & 56.31                         &                                 \\ \cline{3-8}
		                 &                             & Gold-Only SFT & ---              & 59.91            & 60.09           & 65.06           & 64.03                         &                                 \\ \cline{2-9}
		                 & \multirow{4}{*}{Qwen3-8B}
		                 & \textsc{Cls}                & 91.07           & 61.80            & 61.97            & 64.64           & 65.15           & \multirow{4}{*}{$0.121$ n.s.}                                   \\ \cline{3-8}
		                 &                             & Label-Gen SFT & \textbf{92.45}   & 92.88            & 92.45           & 92.19           & 91.93                         &                                 \\ \cline{3-8}
		                 &                             & Label-Gen LogProb & ---              & 56.65            & 56.31           & 57.85           & 58.88                         &                                 \\ \cline{3-8}
		                 &                             & Gold-Only SFT & ---              & 64.55            & 64.64           & 68.50           & 68.84                         &                                 \\
		\hline\hline

		\multirow{16}{*}{SciQ}
		                 & \multirow{4}{*}{Qwen3-0.6B}
		                 & \textsc{Cls}                & \textbf{90.70}  & 68.70            & 68.30            & 72.80           & 70.50           & \multirow{4}{*}{$0.0263^{*}$}                                   \\ \cline{3-8}
		                 &                             & Label-Gen SFT & 88.50            & 86.80            & 87.10           & 86.20           & 84.60                         &                                 \\ \cline{3-8}
		                 &                             & Label-Gen LogProb & ---              & 76.20            & 76.50           & 75.60           & 75.60                         &                                 \\ \cline{3-8}
		                 &                             & Gold-Only SFT & ---              & 84.60            & 85.10           & 85.50           & 86.00                         &                                 \\ \cline{2-9}
		                 & \multirow{4}{*}{Qwen3-1.7B}
		                 & \textsc{Cls}                & \textbf{94.20}  & 83.40            & 83.80            & 85.10           & 84.90           & \multirow{4}{*}{$0.203$ n.s.}                                   \\ \cline{3-8}
		                 &                             & Label-Gen SFT & 93.20            & 91.80            & 91.20           & 90.30           & 89.10                         &                                 \\ \cline{3-8}
		                 &                             & Label-Gen LogProb & ---              & 83.50            & 83.40           & 84.20           & 82.80                         &                                 \\ \cline{3-8}
		                 &                             & Gold-Only SFT & ---              & 91.50            & 91.30           & 92.00           & 92.40                         &                                 \\ \cline{2-9}
		                 & \multirow{4}{*}{Qwen3-4B}
		                 & \textsc{Cls}                & 95.00           & 89.40            & 89.70            & 90.70           & 90.80           & \multirow{4}{*}{$1.000$ n.s.}                                   \\ \cline{3-8}
		                 &                             & Label-Gen SFT & \textbf{95.10}   & 95.20            & 95.20           & 95.30           & 95.20                         &                                 \\ \cline{3-8}
		                 &                             & Label-Gen LogProb & ---              & 90.60            & 90.70           & 90.10           & 90.80                         &                                 \\ \cline{3-8}
		                 &                             & Gold-Only SFT & ---              & 92.60            & 92.90           & 92.70           & 93.00                         &                                 \\ \cline{2-9}
		                 & \multirow{4}{*}{Qwen3-8B}
		                 & \textsc{Cls}                & 96.00           & 93.20            & 93.40            & 93.50           & 94.00           & \multirow{4}{*}{$0.188$ n.s.}                                   \\ \cline{3-8}
		                 &                             & Label-Gen SFT & \textbf{96.90}   & 97.00            & 96.70           & 96.80           & 96.40                         &                                 \\ \cline{3-8}
		                 &                             & Label-Gen LogProb & ---              & 92.20            & 92.60           & 92.50           & 92.20                         &                                 \\ \cline{3-8}
		                 &                             & Gold-Only SFT & ---              & 94.80            & 94.50           & 94.70           & 95.80                         &                                 \\
		\hline
	\end{tabular}
	\caption{Test-set accuracy (\%) on the two science benchmarks
		ARC-Challenge and SciQ. Format and metric definitions are identical
		to Table~\ref{tab:main_results}. \textbf{Bold} marks the highest
		accuracy per (dataset, model) block. Unlike the commonsense benchmarks, the science benchmarks do not show a
		consistent advantage for \textsc{Cls}: Label-Gen SFT directionally wins
		on all ARC-Challenge settings and on the larger SciQ models, while
		\textsc{Cls} is significantly better only on SciQ at 0.6B.}
	\label{tab:science_results}
\end{table}
\clearpage
\FloatBarrier


\end{document}